\documentclass[10pt,conference,pagebackref]{IEEEtran}
\IEEEoverridecommandlockouts
\usepackage{cite}
\usepackage{amsmath,amssymb,amsfonts}
\usepackage{algorithmic}
\usepackage{graphicx}
\usepackage{textcomp}
\usepackage{xcolor}
\usepackage{booktabs} 
\usepackage{bm}
\usepackage{orcidlink}
\usepackage{hyperref}
\usepackage{marvosym}
\hypersetup{
            colorlinks=true,
            linkcolor=red,
            citecolor=blue
            }
\newcommand{\tablestyle}[2]{\setlength{\tabcolsep}{#1}\renewcommand{\arraystretch}{#2}\centering\footnotesize}
\def\BibTeX{{\rm B\kern-.05em{\sc i\kern-.025em b}\kern-.08em
    T\kern-.1667em\lower.7ex\hbox{E}\kern-.125emX}}

\begin{document}
\title{DCAU-Net: Differential Cross Attention and Channel-Spatial Feature Fusion for Medical Image Segmentation
\thanks{This work described in this paper was supported in part by the Scientific Research Foundation of Chongqing University of Technology under Grant 2021ZDZ030 and in part by the Youth Project of Science and Technology Research Program of Chongqing Education Commission of China under Grant KJQN202301145.}
}

\author{\IEEEauthorblockN{Yanxin Li$^{\orcidlink{0009-0009-7784-7908}}$, Hui Wan$^{\orcidlink{0009-0007-1867-3747}}$, and Libin Lan$^{\orcidlink{0000-0003-4754-813X}}$$(\textrm{\Letter})$}
\IEEEauthorblockA{College of Computer Science and Engineering, 
Chongqing University of Technology,
Chongqing, China \\
\{yanxin.li, huiwan\}@stu.cqut.edu.cn, lanlbn@cqut.edu.cn}
}

\maketitle

\begin{abstract}
Accurate medical image segmentation requires effective modeling of both long-range dependencies and fine-grained boundary details. 
While transformers mitigate the issue of insufficient semantic information arising from the limited receptive field inherent in convolutional neural networks, they introduce new challenges: standard self-attention incurs quadratic computational complexity and often assigns non-negligible attention weights to irrelevant regions, diluting focus on discriminative structures and ultimately compromising segmentation accuracy. Existing attention variants, although effective in reducing computational complexity, fail to suppress redundant computation and inadvertently impair global context modeling.
Furthermore, conventional fusion strategies in encoder-decoder architectures, typically based on simple concatenation or summation, can not adaptively integrate high-level semantic information with low-level spatial details. To address these limitations, we propose DCAU-Net, a novel yet efficient segmentation framework with two key ideas. First, a new Differential Cross Attention (DCA) is designed to compute the difference between two independent softmax attention maps to adaptively highlight discriminative structures. By replacing pixel-wise key and value tokens with window-level summary tokens, DCA dramatically reduces computational complexity without sacrificing precision. Second, a Channel-Spatial Feature Fusion (CSFF) strategy is introduced to adaptively recalibrate features from skip connections and up-sampling paths through using sequential channel and spatial attention, effectively suppressing redundant information and amplifying salient cues.
Experiments on two public benchmarks demonstrate that DCAU-Net achieves competitive performance with enhanced segmentation accuracy and robustness.
\end{abstract}

\begin{IEEEkeywords}
Convolutional neural network, differential cross attention, feature fusion, medical image segmentation, transformer.
\end{IEEEkeywords}

\section{Introduction}
Accurate medical image segmentation is a cornerstone of computer-aided diagnosis and clinical decision-making. Precise delineation of anatomical structures and pathological regions across diverse imaging modalities, such as magnetic resonance imaging (MRI) and computed tomography (CT), enables critical applications, including disease diagnosis, treatment planning, and image-guided surgery~\cite{b1},~\cite{b2},~\cite{b3}.

Early medical image segmentation models, such as U-Net~\cite{b4} and its numerous variants~\cite{b5},~\cite{b6},~\cite{b7}, are predominantly built upon convolutional neural networks (CNNs), which leverage local receptive fields and parameter sharing for efficient feature extraction. However, the inherent local inductive bias of CNNs limits their ability to model long-range dependencies critical for accurately modeling global anatomical context in complex medical images.

Recently, transformers have been introduced into medical image segmentation to address the limitations of CNNs in capturing long-range dependencies~\cite{b1},~\cite{b8}. By leveraging self-attention, these models explicitly capture pixel-wise global contextual relationships, significantly enhancing semantic consistency across the entire image. However, standard self-attention adopts a pixel-wise query-key-value computation paradigm, resulting in a high computational complexity of $\mathcal{O}(N^2)$~\cite{b9}. Moreover, it tends to assign non-negligible attention weights to irrelevant or redundant regions, thereby weakening the focus on discriminative features~\cite{b10}.

To reduce such computational cost, a series of efficient attention variants have been introduced into medical image segmentation field,
including window-based attention~\cite{b2},~\cite{b12},~\cite{b13}, axial attention~\cite{b14},~\cite{b15}, and dynamic sparse attention~\cite{b16}. Nevertheless, they still have inherent limitations. For instance, window-based attention is confined to local interactions, and even with shift strategies, it falls short of achieving true global modeling; axial attention splits spatial dimensions, disrupting the holistic correlation of features; and dynamic sparse attention suffers from training instability due to dynamic sparsity and suboptimal hardware utilization compared to static attention. 

Consequently, existing efficient attention primarily prioritizes computational efficiency but often reintroduces local inductive biases, typically by partitioning spatial regions, which undermines their capacity to model long-range dependencies effectively. This raises a critical challenge: how to simultaneously achieve low computational cost and high-quality, globally aware attention modeling in medical image segmentation.

In addition, within the encoder--decoder architectures, the decoder commonly fuses features from skip connections and upsampled decoder features through simple operations such as concatenation or element-wise addition~\cite{b1},~\cite{b4}. Such fusion strategies struggle to suppress redundant features and fail to fully exploit the complementary information between high-level semantic content and low-level spatial details.
Although channel and spatial attention mechanisms have been widely adopted for feature recalibration in general computer vision tasks~\cite{b17},~\cite{b18},~\cite{b19}, 
to the best of our knowledge, they have not yet been explored in medical image segmentation for adaptive fusion of features from skip connections and up-sampling paths. This highlights the need for a fusion strategy that can jointly enhance discriminative features and suppress redundant ones across both channel and spatial dimensions.

In this paper, we propose a lightweight yet efficient medical image segmentation framework, termed DCAU-Net. The DCAU-Net is built upon two key ideas.
First, it adapts the differential attention~\cite{b10}, originally proposed for natural language processing under a ``pixel-wise query–key–value’’ self-attention paradigm, to the medical vision domain by reformulating it into a ``pixel-wise query -- window-level key-value’’ cross-attention paradigm, which is referred to as \textbf{D}ifferential \textbf{C}ross \textbf{A}ttention (\textbf{DCA}). This design computes the difference between two independent softmax attention maps to suppress noise and enhance focus on relevant structures. By summarizing keys and values at the window level, it reduces computational complexity by a factor of $M^2$ (where $M$ denotes the window size), while effectively suppressing background artifacts, avoiding distraction from irrelevant regions, and preserving fine-grained boundary sensitivity.
Second, DCAU-Net introduces the \textbf{C}hannel-\textbf{S}patial \textbf{F}eature \textbf{F}usion (\textbf{CSFF}) strategy to better integrate encoder features via skip connections and decoder features through up-sampling paths, while suppressing redundancy and amplifying discriminative cues via lightweight convolutions followed by sequential channel and spatial attention.

Our main contributions are as follows: 

(1) We propose a novel Differential Cross Attention (DCA) mechanism that adapts differential attention by replacing pixel-wise key and value tokens with window-level summary tokens, enabling computationally efficient yet high-quality attention modeling; 

(2) We introduce a Channel-Spatial Feature Fusion (CSFF) strategy, which adaptively recalibrates the features from skip connections and up-sampling paths to suppress redundancy and enhance discriminative cues; and

(3) We integrate DCA and CSFF into a unified U-shaped framework, DCAU-Net, achieving competitive segmentation performance on two public benchmarks. Code will be released upon acceptance.

\section{Method}
\subsection{Differential Cross Attention}
\begin{figure}[htbp]
\centering
\includegraphics[width=1.0\columnwidth]{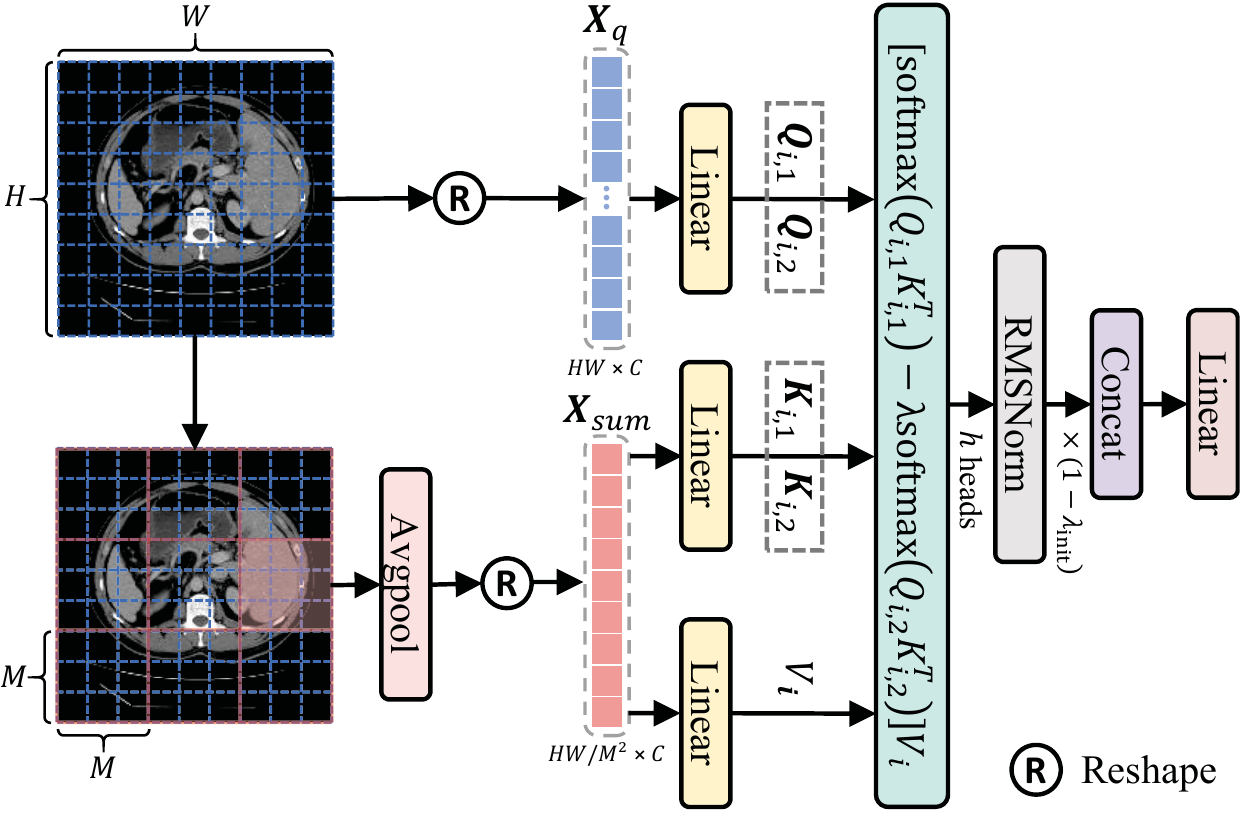}
\caption{Details of the differential cross attention. It performs efficient cross attention between pixel-wise queries and window-level key–value pairs via differential attention, suppressing redundancy and enhancing focus on discriminative structures.}
\label{fig1:DCA}
\end{figure}

As shown in Fig.~\ref{fig1:DCA}, the DCA adapts the differential attention~\cite{b10} to an efficient cross attention framework. Given an input feature map $\bm{X} \in \mathbb{R}^{H \times W \times C}$, pixel-wise query tokens $\bm{X}_q \in \mathbb{R}^{N \times C}$ with $N = HW$ are obtained by flattening $\bm{X}$. Meanwhile, window-level summary tokens $\bm{X}_{sum} \in \mathbb{R}^{N_{win} \times C}$, where $N_{win} = HW/M^2$, are generated by partitioning $\bm{X}$ into non-overlapping $M \times M$ spatial windows (default $M=$ 7) and applying average pooling within each window. This design enables fine-to-coarse interaction while reducing computational cost.

We then compute cross attention between the pixel-wise query tokens $\bm{X}_q$ and the window-level summary tokens $\bm{X}_{sum}$ using a multi-head differential attention mechanism, denoted as MHDiffAttn. The output of the DCA is defined as:
\begin{equation}
    \text{DCA}(\bm{X}) = \text{MHDiffAtt}(\bm{X}_q, \bm{X}_{sum}, \bm{X}_{sum}).
\end{equation}

For each head $i \in \{1, \dots, h\}$, we project $\bm{X}_q$ and $\bm{X}_{sum}$ into queries, keys, and values:
$\bm{Q}_{i,1}, \bm{Q}_{i,2} \in \mathbb{R}^{N \times d}$, 
$\bm{K}_{i,1}, \bm{K}_{i,2} \in \mathbb{R}^{N_{win} \times d}$, 
and $\bm{V}_i \in \mathbb{R}^{N_{win} \times 2d}$. 
The projections are defined as:
\begin{equation}
\begin{split}
    [\bm{Q}_{i,1}; \bm{Q}_{i,2}] &= \bm{X}_q \bm{W}^Q_i,\\
    [\bm{K}_{i,1}; \bm{K}_{i,2}] &= \bm{X}_{sum} \bm{W}^K_i, \\
    \bm{V}_i &= \bm{X}_{sum} \bm{W}^V_i,
\end{split}
\end{equation}
where $\bm{W}_i^Q, \bm{W}^K_i, \bm{W}^V_i \in \mathbb{R}^{C \times 2d}$ are learnable projection matrices, and 2$d$ denotes the dimension of each head.

For $\bm{Q}$ and $\bm{K}$, we compute two independent softmax attention maps:
\begin{equation}
    \bm{S}_{i,1} = \operatorname{softmax}\left( \frac{\bm{Q}_{i,1} \bm{K}_{i,1}^\top}{\sqrt{d}} \right), \bm{S}_{i,2} = \operatorname{softmax}\left( \frac{\bm{Q}_{i,2} \bm{K}_{i,2}^\top}{\sqrt{d}} \right),
\end{equation}
and form the output of the $i$-th attention head as:
\begin{equation}
    \text{head}_i = \left( \bm{S}_{i,1} - \lambda \bm{S}_{i,2} \right) \bm{V}_i,
\end{equation}
where the $\lambda$ is a learnable scalar shared across all heads within the same layer and re-parameterized as:
\begin{equation}
    \lambda = \exp(\bm{\lambda}_{q_1} \cdot \bm{\lambda}_{k_1}) - \exp(\bm{\lambda}_{q_2} \cdot \bm{\lambda}_{k_2}) + \lambda_{init},
\end{equation}
with learnable vectors $\bm{\lambda}_{q_1}, \bm{\lambda}_{k_1}, \bm{\lambda}_{q_2}, \bm{\lambda}_{k_2} \in \mathbb{R}^{d}$ and a depth-dependent initialization $\lambda_{init} = 0.8 - 0.6 \exp(-0.3 \cdot (l-1))$, 
where $l$ is the index of the DCA block.
Each head output is then normalized and scaled independently:
\begin{equation}
    \widetilde{\text{head}}_i = (1 - \lambda_{init}) \cdot \operatorname{RMSNorm}(\text{head}_i).
\end{equation}
Here, RMSNorm~\cite{zhang-sennrich-neurips19} is used instead of Layer Normalization to reduce computational overhead.

Finally, all head outputs are concatenated and projected back to the model dimension via a learnable matrix $\bm{W}_O \in \mathbb{R}^{C \times C}$:
\begin{equation}
     \text{MHDiffAttn}(\bm{X}) = \operatorname{Concat}(\widetilde{\text{head}}_1, \dots, \widetilde{\text{head}}_h) \bm{W}_O,
\end{equation}
where the number of heads is set to $h = C / (2d)$.

Notably, when $M=H=W$, the summary tokens collapse to a single global token, and the DCA module reverts to standard differential attention instead of performing differential cross attention.

\subsection{Differential Cross Attention Block}
\begin{figure}[b]
\centering
\includegraphics[width=1.0\columnwidth]{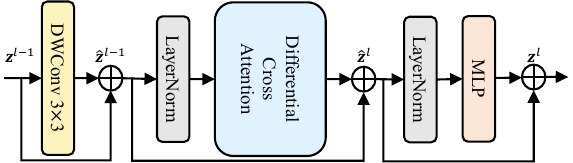}
\caption{Details of the DCA Block, consisting of a 3$\times$3 depth-wise convolution, a DCA module, and a 2-layer MLP.}
\label{fig2:dca_block}
\end{figure}

Building on the above proposed DCA mechanism, we design a DCA block, as illustrated in Fig.~\ref{fig2:dca_block}.
The DCA block comprises three main components: a 3$\times$3 depth-wise convolution, a DCA module, and a 2-layer MLP with an expansion ratio of $e=$ 4. Each component is equipped with a residual connection, followed by Layer Normalization. Formally, the DCA Block is computed as follows:
\begin{align}
    \hat{\bm{z}}^{l-1} &= f_{\text{dwc}}^{3\times3}(\bm{z}^{l-1}) + \bm{z}^{l-1}, \\
    \hat{\bm{z}}^{l}   &= \text{DCA}\big( \operatorname{LayerNorm}(\hat{\bm{z}}^{l-1}) \big) + \hat{\bm{z}}^{l-1},  \\
    \bm{z}^{l}         &= \text{MLP}\big( \operatorname{LayerNorm}(\hat{\bm{z}}^{l}) \big) + \hat{\bm{z}}^{l},
\end{align}
where $f_{\text{dwc}}^{3\times3}$ denotes a 3$\times$3 depth-wise convolution, and $\hat{\bm{z}}^{l-1}$, $\hat{\bm{z}}^{l}$, $\bm{z}^{l}$ denote the intermediate outputs by the depth-wise convolution, DCA module, and MLP, respectively.

\subsection{Channel-Spatial Feature Fusion Block}
As previously mentioned, the primary goal of the Channel-Spatial Feature Fusion (CSFF) strategy is to adaptively recalibrate features from skip connections and up-sampling paths so as to suppress redundancy and enhance discriminative cues. 
To this end, we design a CSFF block, as shown in Fig.~\ref{fig3:net} (right).
The CSFF block fuses encoder and decoder features $\bm{X}_e, \bm{X}_d \in \mathbb{R}^{H \times W \times C}$ by first refining each input separately and then combining them into an intermediate representation $\bm{X}_f \in \mathbb{R}^{H \times W \times C}$:
\begin{align}
    \tilde{\bm{X}}_e &= \gamma \big( \mathrm{BN}( f_{3\times3}(\bm{X}_e) ) \big), \\
    \tilde{\bm{X}}_d &= \gamma \big( \mathrm{BN}( f_{3\times3}(\bm{X}_d) ) \big), \\
    \bm{X}_f &= \gamma \big( \mathrm{BN}( f_{3\times3}( \mathrm{Concat}( \tilde{\bm{X}}_e, \tilde{\bm{X}}_d ) ) ) \big),
\end{align}
where $f_{k\times k}$ denotes a $k\times k$ convolution, $\gamma(\cdot)$ is ReLU, $\mathrm{BN}(\cdot)$ is batch normalization, and $\mathrm{Concat}(\cdot)$ denotes channel-wise concatenation.
Since the resulting $\bm{X}_f$ exhibits redundancy in both channel and spatial dimensions, we sequentially apply channel and spatial attention mechanisms to selectively enhance informative features.

First, the channel attention learns a channel-wise weight map from global spatial statistics:
\begin{align}
    \bm{F}^c_{\mathrm{avg}} &= \mathrm{AvgPool}(\bm{X}_f), \quad
    \bm{F}^c_{\mathrm{max}} = \mathrm{MaxPool}(\bm{X}_f), \\
    \bm{M}_c &= \sigma \big( \mathrm{MLP}(\bm{F}^c_{\mathrm{avg}}) + \mathrm{MLP}(\bm{F}^c_{\mathrm{max}}) \big), \\
    \bm{X}_c &= \bm{M}_c \odot \bm{X}_f,
\end{align}
where $\mathrm{AvgPool}$ and $\mathrm{MaxPool}$ operate over the spatial dimensions, $\mathrm{MLP}$ denotes a shared 2-layer perceptron with reduction ratio $r=4$, $\sigma(\cdot)$ is the sigmoid activation, and $\odot$ denotes element-wise multiplication.

Second, the spatial attention produces a spatial weight map from channel-aggregated features:
\begin{align}
    \bm{F}^s_{\mathrm{avg}} &= \mathrm{AvgPool}(\bm{X}_c), \quad
    \bm{F}^s_{\mathrm{max}} = \mathrm{MaxPool}(\bm{X}_c), \\
    \bm{M}_s &= \sigma \left( f_{3\times3} \left( \mathrm{Concat} \left( \bm{F}^s_{\mathrm{avg}},\, \bm{F}^s_{\mathrm{max}} \right) \right) \right), \\
    \bm{X}_o &= \bm{M}_s \odot \bm{X}_c,
\end{align}
where $\mathrm{AvgPool}$ and $\mathrm{MaxPool}$ operate along the channel dimension, producing two maps of size $H$$\times$$W$$\times$$1$.

\subsection{Overall Architecture}
\begin{figure*}[htbp]
\centering
\includegraphics[width=1.0\textwidth]{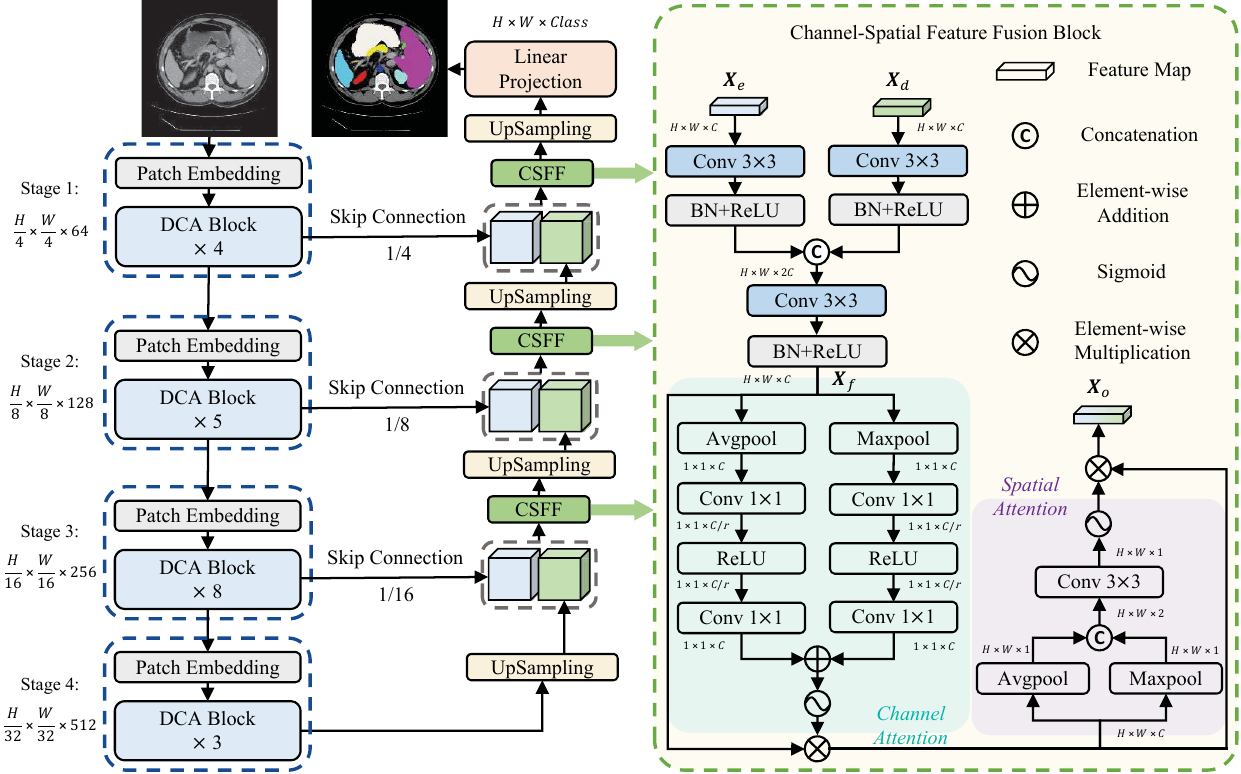}
\caption{Overall architecture of the proposed DCAU-Net. The network adopts a U-shaped encoder-decoder framework with four hierarchical stages. The encoder integrates DCA blocks, centered on the differential cross attention. Features from the encoder are transferred to the decoder via skip connections and adaptively fused with those from previous decoder layers through CSFF blocks to enhance segmentation accuracy.}
\label{fig3:net}
\end{figure*}

The overall architecture of the proposed DCAU-Net is illustrated in Fig.~\ref{fig3:net}. DCAU-Net adopts a U-shaped structure commonly used in segmentation tasks and consists of three main components:
a DCA block-based encoder, skip connections, and a CSFF block-based decoder. 
The encoder adopts a four-stage hierarchical structure, where spatial downsampling is achieved through Patch Embedding Layers, progressively reducing the feature map resolution while increasing the channel dimension. Each stage employs the Differential Cross Attention (DCA) block, whose core component is the DCA module that leverages window-level summary tokens derived via average pooling to compute keys and values. 
This allows for capturing long-range dependencies through cross attention. 
The decoder symmetrically performs four upsampling steps to restore the original spatial resolution and fuses the features from corresponding encoder stages and from the previous decoder layer. 
To enable more refined encoder-decoder feature fusion, we introduce the Channel-Spatial Feature Fusion (CSFF) block into the decoder, which conducts dual adaptive calibration in both channel and spatial dimensions. 
Finally, the network outputs a pixel-wise segmentation mask with the same resolution as the input image.

\section{Experiments}
\subsection{Datasets}
\noindent\textbf{Synapse Multi-Organ Segmentation Dataset}~\cite{Landman}.
The dataset originates from the MICCAI 2015 Multi-Atlas Abdomen Labeling Challenge and consists of 30 abdominal contrast-enhanced CT scans. It is split into 18 scans for the training set and 12 for the testing set. Each scan includes expert-annotated segmentations for eight abdominal organs: aorta, gallbladder, liver, spleen, stomach, pancreas, left kidney, and right kidney. Performance is evaluated using the average Dice Similarity Coefficient (DSC) and the average Hausdorff Distance (HD) across all target structures.

\noindent\textbf{Automated Cardiac Diagnosis Challenge (ACDC) Dataset}~\cite{Bernard2018ACDC}.
The ACDC dataset contains cardiac MRI scans acquired from 100 patients. Each scan is accompanied by expert-provided ground-truth segmentations for three key anatomical structures: left ventricle (LV), right ventricle (RV), and myocardium (MYO). The scans are allocated into 70 for the training set, 10 for the validation set, and 20 for the testing set. Following~\cite{b1,b2}, evaluation is based solely on the average DSC across the three structures.

\subsection{Implementation Details}
The proposed DCAU-Net is implemented in PyTorch 2.0 and trained on an NVIDIA GeForce RTX 3090 GPU. The encoder backbone is initialized with weights pretrained on ImageNet. For both the Synapse and ACDC datasets, input images are resized to 224$\times$224, and standard data augmentations including random rotation and flipping are applied during training. The model is trained for 400 epochs with a batch size of 24, optimized using the AdamW optimizer with a weight decay of 1$e-$4. The initial learning rate is set to 1$e-$3 for Synapse and 3$e-$4 for ACDC, optimizing a combined Cross-entropy (0.4) and DICE (0.6) loss.

\begin{table*}[htbp]
\centering
% \footnotesize
\caption{Quantitative results of our approach against other state-of-the-art methods on the Synapse dataset. The symbol $\uparrow$ indicates the larger the better. The symbol $\downarrow$ denotes the smaller the better. The best result is in \textbf{blod}, and the second best is \underline{underlined}.}
\resizebox{1.0\linewidth}{!}{%
\begin{tabular}{l|cc|cc|cccccccc}
\toprule
Methods & Params (M) &Flops (G) & DSC (\%) $\uparrow$  & HD (mm) $\downarrow$  & Aorta &Gallbladder & Kidney (L)  & Kidney (R)  & Liver  & Pancreas & Spleen  & Stomach\\
\midrule
U-Net~\cite{b4} &14.80 &8.43   &76.85   &39.70 &\underline{89.07} &69.72 &77.77 &68.60 &93.43 &53.98 &86.67 &75.58 \\
Att-UNet~\cite{b6} &34.88 &66.64  &77.77  &36.02 &\textbf{89.55} &68.88 &77.98 &71.11 &93.57 &58.04 &87.30 &75.75 \\
TransUNet~\cite{b1} &105.28 &29.35 &77.48   &31.69 &87.23 &63.13 &81.87 &77.02 &94.08 &55.86 &85.08 &75.62 \\
Swin-Unet~\cite{b2} &27.17 &6.16 &79.13  &21.55 & 85.47 &66.53  & 83.28 &79.61  &94.29 & 56.58 &90.66  & 76.60 \\
HiFormer~\cite{b13} &25.51 &8.05 &80.39  &\textbf{14.70} & 86.21 &65.69  & 85.23 &79.77  &94.61 & 59.52 &90.99 & 81.08 \\
PVT-CASCADE~\cite{b20} &35.28 &7.62 &81.06 &20.23& 83.01 &\underline{70.59} & 82.23 &80.37  &94.08 & 64.43 &90.10  & \textbf{83.69} \\
MISSFormer~\cite{b11} &42.46 &9.89 &81.96  &18.20& 86.99 &68.65  &85.21  &\underline{82.00}  &94.41 & \textbf{65.67} &\underline{91.92} & 80.81 \\
BRAU-Net++~\cite{b16} &62.63 &17.66 &\underline{82.47}  &19.07 & 87.95 &69.10  &\underline{87.13} &81.53 &\underline{94.71}  &\underline{65.17} &91.89 &\underline{82.26}\\
\midrule
DCAU-Net (Ours) &21.56 &4.67 &\textbf{83.29} & \underline{15.14} & 87.91 & \textbf{73.09} & \textbf{88.20} & \textbf{84.05} & \textbf{94.98} & 63.89 & \textbf{93.26} & 80.94 \\
\bottomrule
\end{tabular} }
\label{tab1:synapse}
\end{table*}

\begin{figure*}[htbp]
\centering
\includegraphics[width=1.0\linewidth, keepaspectratio]{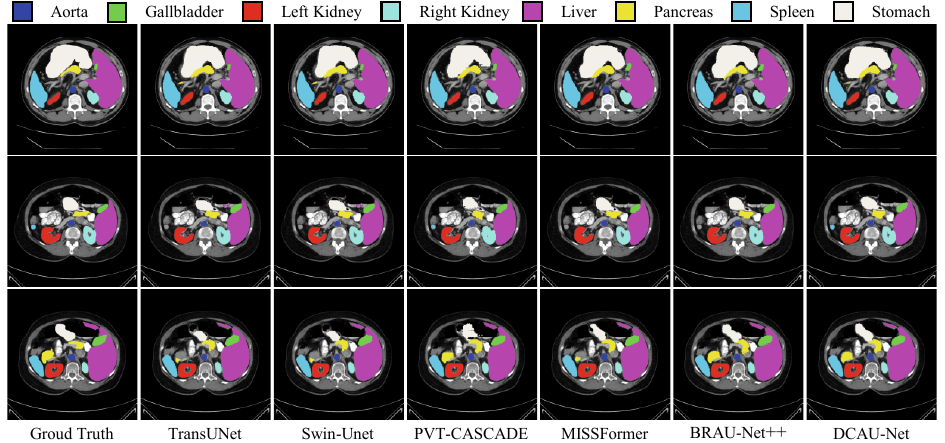}
\caption{Qualitative comparisons of our approach against other state-of-the-art methods on the Synapse dataset.}
\label{synapse_results}
\end{figure*}

\begin{table}[htbp]
\centering
\scriptsize
% \tiny
\caption{Quantitative results of different methods on ACDC dataset.}
\label{tab2:acdc}
\begin{tabular}{l|c|ccc}
\toprule
Method & DSC (\%) $\uparrow$ & RV & Myo & LV \\
\midrule
R50+U-Net~\cite{b1}      & 87.55 & 87.10 & 80.63 & 94.92 \\
R50+Att-UNet~\cite{b1}   & 86.75 & 87.58 & 79.20 & 93.47 \\
UNETR~\cite{b3} &88.61 &85.29 &86.52 &94.02\\
TransUNet~\cite{b1}      & 89.71 & 88.86 & 84.54 & 95.73 \\
Swin-Unet~\cite{b2}            & 90.00 & 88.55 & 85.62 & 95.83 \\
MISSFormer~\cite{b11}   & 90.86 & 89.55 & 88.04 & 94.99 \\
PVT-CASCADE~\cite{b20}    & 91.46 & 88.90 & \textbf{89.97} & 95.50 \\
BRAU-Net++~\cite{b16}           & \underline{92.07} & \textbf{90.72} & 89.57 & \underline{95.90} \\
\midrule
DCAU-Net (Ours)                        & \textbf{92.11} & \underline{90.46} & \underline{89.88} & \textbf{95.98} \\
\bottomrule
\end{tabular}
\end{table}

\begin{table*}[htbp]
\centering
\scriptsize
\tablestyle{9pt}{1.1}
\caption{Ablation study of pre-trained weights for DCAU-Net on the Synapse dataset.}
\label{tab3:ablation_pre}
\resizebox{1.0\linewidth}{!}{%
\begin{tabular}{c|cc|cccccccc}
\toprule
Pre-trained & DSC (\%) $\uparrow$ & HD (mm) $\downarrow$ & Aorta & Gallbladder & Kidney (L) & Kidney (R) & Liver & Pancreas & Spleen & Stomach \\
\midrule
w/o & 81.25 & 17.63 & 87.68 & 67.88 & 86.77 & 82.31 & 94.36 & 59.56 & 92.02 & 79.43 \\
w  & \textbf{83.29} & \textbf{15.14} & \textbf{87.91} & \textbf{73.09} & \textbf{88.20} & \textbf{84.05} & \textbf{94.98} & \textbf{63.89} & \textbf{93.26} & \textbf{80.94} \\
\bottomrule
\end{tabular}
}
\end{table*}

\subsection{Comparison on Synapse Dataset}
We evaluate DCAU-Net on the Synapse dataset against recent CNN, Transformer, and hybrid-based methods. As shown in Table~\ref{tab1:synapse}, our method achieves a new state-of-the-art DSC of 83.29\% with only 4.67G FLOPs -- the lowest among all competitive approaches -- and just 21.56M parameters. It also attains the second-best HD of 15.14 mm, demonstrating strong boundary accuracy. Notably, DCAU-Net yields the highest organ-wise DSC on the gallbladder (73.09\%), left kidney (88.20\%), right kidney (84.05\%), liver (94.98\%), and spleen (93.26\%), highlighting its robustness on both small and complex anatomical structures.

Qualitative comparisons in Fig.~\ref{synapse_results} further illustrate that DCAU-Net produces more accurate masks, especially for challenging organs like the gallbladder and spleen.

\subsection{Comparison on ACDC Dataset}
We further evaluate DCAU-Net on the ACDC dataset. As shown in Table~\ref{tab2:acdc}, our method achieves a new state-of-the-art overall DSC of 92.11\%. It delivers the best performance on both Myo and LV, demonstrating superior accuracy in segmenting clinically critical cardiac structures.

\subsection{Ablation Study}
\subsubsection{Effectiveness of Pre-trained Weights}
We evaluate the impact of ImageNet pre-trained weights on DCAU-Net using the Synapse dataset. As shown in Table~\ref{tab3:ablation_pre}, training from random initialization achieves a DSC of 81.25\% and HD of 17.63 mm. Initializing with pre-trained weights improves DSC by 2.04\% and reduces HD by 2.49 mm, demonstrating the benefit of pre-trained weights. For consistency and to isolate architectural contributions, all subsequent ablation studies use random initialization.

\subsubsection{Effectiveness of Different Attention and $\lambda$ Initialization Strategy in DCA}
Table~\ref{tab4:ablation_dca} shows the impact of different attention and $\lambda$ initialization strategies in the DCA under the ``pixel-wise query -- window-level key-value'' cross-attention paradigm. The baseline employs standard scaled dot-product attention~\cite{b9}. We compare it against differential attention with fixed $\lambda_{init} \in \{0.5, 0.8\}$ and the dynamic  initialization schedule $\lambda_{init} = 0.8 - 0.6\exp(-0.3 \cdot (l - 1))$~\cite{b10}. Results show that differential attention consistently outperforms the baseline, with the dynamic initialization achieving the best performance, demonstrating that our efficient cross attention design effectively preserves the capacity to adaptively suppress redundant attention computation.

\begin{table}[htbp]
\centering
\footnotesize
\caption{Ablation study of the attention type and $\lambda$ initialization strategy for DCA on the Synapse dataset.}
\label{tab4:ablation_dca}
\begin{tabular}{l|c|cc}
\toprule
Attention Type & $\lambda$ Init Strategy & DSC (\%) $\uparrow$ & HD (mm) $\downarrow$ \\
\midrule
Standard Attention          & -                       & 80.79 & 19.35 \\
Differential Attention       & Fixed (0.5)             & 81.18 & 19.10 \\
Differential Attention       & Fixed (0.8)             & 81.03 & 17.94 \\
Differential Attention & Dynamic & \textbf{81.25} & \textbf{17.63} \\
\bottomrule
\end{tabular}
\end{table}

\subsubsection{Effectiveness of CSFF}
We conduct an ablation study to evaluate the CSFF block by comparing four decoder fusion strategies: standard U-Net fusion (baseline), CSFF without spatial attention (w/o SA), CSFF without channel attention (w/o CA), and the full CSFF.
As shown in Table~\ref{tab5:ablation_csff}, the full CSFF outperforms the baseline by +1.49\% in DSC and reduces HD by 2.87 mm. Removing either attention component degrades performance, demonstrating that channel and spatial attention provide complementary benefits and their joint integration is essential for effective feature fusion.

\begin{table}[htbp]
\centering
\caption{Ablation study of the CSFF block on the Synapse dataset.}
\label{tab5:ablation_csff}
\begin{tabular}{l|cc}
\toprule
Fusion Strategy       & DSC (\%) $\uparrow$ & HD (mm) $\downarrow$ \\
\midrule
Standard U-Net Fusion & 79.76    & 20.50 \\
CSFF (w/o SA)         & 79.63    & 23.19 \\
CSFF (w/o CA)         & 80.83    & 21.56 \\
CSFF (Full)  & \textbf{81.25} & \textbf{17.63} \\
\bottomrule
\end{tabular}
\end{table}

\section{Conclusion}
In this paper, we propose DCAU-Net, a lightweight yet powerful medical image segmentation framework with two key designs: (1) a differential cross attention mechanism that efficiently applies differential attention between pixel-wise queries and window-level key–value pairs; and (2) a channel-spatial feature fusion strategy that adaptively integrates encoder and decoder features through joint channel and spatial attention. DCAU-Net achieves state-of-the-art performance on the Synapse and ACDC datasets, demonstrating superior accuracy with lower computational cost.

% \section*{Acknowledgment}
% This work described in this paper was supported in part by the Scientific Research Foundation of Chongqing University of Technology under Grant 2021ZDZ030 and in part by the Youth Project of Science and Technology Research Program of Chongqing Education Commission of China under Grant KJQN202301145.

\bibliographystyle{IEEEtran}
\bibliography{references} 

\end{document}